\documentclass[11pt,a4paper]{article}

\usepackage{booktabs} 
\usepackage{amsmath}
\usepackage{amssymb} 
\usepackage{subcaption} 
\usepackage{graphicx}
\usepackage{tikz}
\usepackage{pgfplots}
\pgfplotsset{compat=1.18}
\usetikzlibrary{patterns}
\usepackage{placeins}

\usepackage[nocustomfonts]{nxai}
\newcommand{\affmark}[1]{\raisebox{0.55ex}{\fontsize{7}{7}\selectfont #1}}

\nxaisetalogo{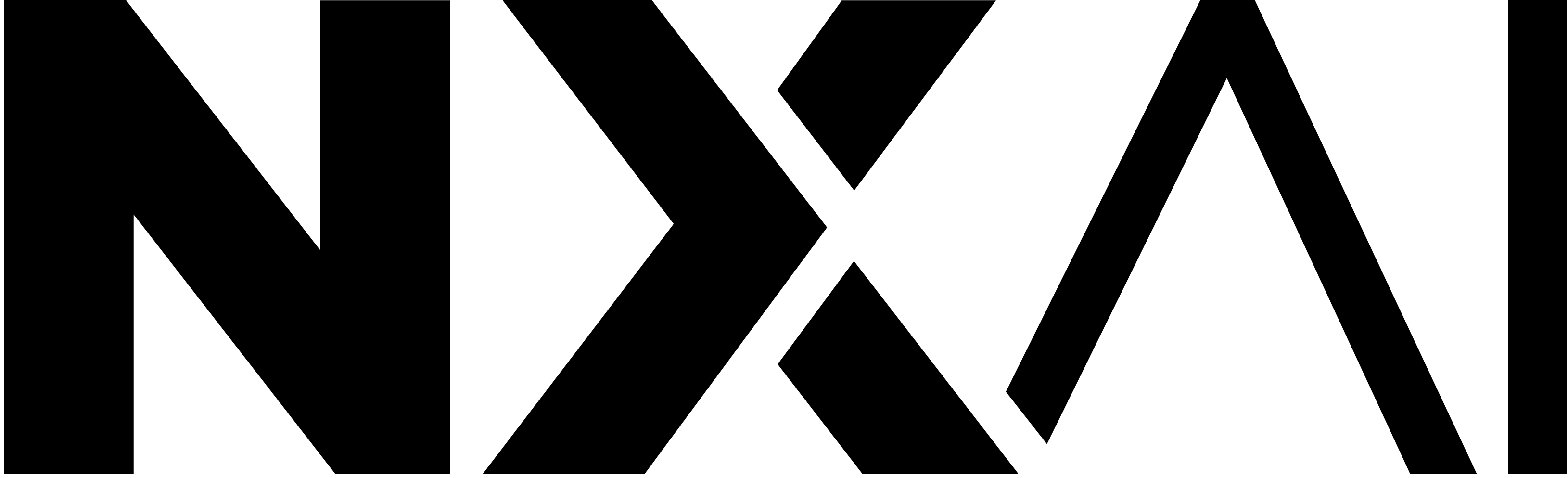} 
\nxaisetlogoheight{4mm} 
\nxaisetdate{\today}
\nxaisetfooter{Corresponding author: Lucia Licakova (lucia.licakova@gmail.com).}

\title{Architectures and KV Replay for Depth-Recurrent Language Models}
\author{Lucia Licakova\affmark{1,2} \and Niklas Schmidinger\affmark{1,2} \and Lukas Hauzenberger\affmark{1,2} \and Sebastian Böck\affmark{1} \and Sepp Hochreiter\affmark{1,2}}
\nxaisetaffiliations{%
\affmark{1}NXAI, \affmark{2}Johannes Kepler University Linz, Austria.
}

\nxaiabstract{%
Depth recurrence increases effective model depth by repeatedly applying a fixed computation block. However, sharing attention parameters does not share the key--value (KV) states generated at different recurrence iterations, which causes memory growth with recurrence depth during autoregressive decoding. We study how to reduce this memory cost through architectural and cache-sharing approaches within a common depth-recurrent framework. We first study a looped hybrid that repeats a stack of Transformer layers, with recurrent mLSTM or Gated DeltaNet blocks replacing most attention layers and periodic Transformer attention anchors retained. We then restrict recurrence to a dedicated recurrent workspace in a \textsc{Parcae}-style architecture and test its stabilising adapter as a wrapper for the looped hybrid. Finally, we train a looped Transformer with recurrent KV replay, where only selected recurrence iterations write KV states that later iterations reuse. This trains the model to operate with the same shared KV topology used during inference. We find that looped hybrids require sufficiently frequent attention anchors, with performance depending strongly on the recurrent mixer. For cross-iteration KV sharing, recurrent KV replay allows multiple recurrence iterations to reuse the same KV states while retaining performance close to the full looped Transformer with separate KV caches for each iteration.
}

\begin{document}
\maketitle

\section{Introduction}
Depth-recurrent language models increase computational depth without allocating a distinct set of parameters to every effective layer. A stack of $L$ physical blocks applied for $R$ recurrence iterations implements an effective depth of $LR$ while retaining substantially fewer parameters than an independently parameterized $LR$-layer Transformer. This construction is attractive when parameter storage or optimiser state is the principal constraint. It also permits inference-time compute scaling by changing the number of recurrence iterations without changing the stored model.
However, autoregressive decoding stores keys and values whose size grows with sequence length, layer count, batch size, head dimension, and, in a looped architecture, the number of recurrence iterations whose attention states must remain distinct. Parameter sharing across depth does not imply memory sharing across depth, and a question arises:
\begin{nxaiinfo}[Research question]
How can inference memory be reduced in depth-recurrent language models while retaining the benefits of exact attention?
\end{nxaiinfo}
To study this question, we make three empirical contributions. First, we evaluate a looped hybrid architecture that replaces most repeated attention layers with recurrent sequence mixers while retaining periodic Transformer~\citep{vaswani2017attention} attention anchors. Second, we study a \textsc{Parcae}-style architecture~\citep{prairie2026parcae}, which confines recurrence to a dedicated latent workspace between Transformer encoder and decoder stages, and assess whether its stabilising adapter transfers to the looped-hybrid architecture. Third, we introduce cache-compression training, which trains a looped Transformer under the same KV-sharing policy that will later be used during inference.
The first two contributions reduce inference memory through architectural changes. The third one leaves the architecture unchanged but reduces the number of persistent KV groups through training.

\section{Related Work}
\paragraph{Looped attention.} Recent work studies how to make recurrence efficient by changing the token mixer itself. LT2~\citep{deng2026lt2} compares full, linear, sparse, and hybrid attention operators inside recurrent loops. \citet{li2025skip} show that pretrained Transformers can also adapt computation at inference time by dynamically skipping or repeating layers. \textsc{Parcae}~\citep{prairie2026parcae}, separates computation into prelude, core, and coda stages. It confines recurrence to the latent core and applies a stabilising adapter that repeatedly injects the original representation into the recurrent state. 
\paragraph{Layer-wise and cross-layer KV sharing.}
Cross-Layer Attention and related methods share key--value representations across different Transformer layers~\citep{brandon2024crosslayer,wu2025systematic,yang2024kvsharer}. In contrast, we share KV states across recurrence iterations of the same weight-shared attention layer. Mixture-of-Recursions~\citep{mor2025} routes tokens through different numbers of shared-layer recursions and includes a variant that reuses KV pairs from the first recursion. Parallel Loop Transformer~\citep{wu2025plt} shares the first loop's KV cache across subsequent loops and combines it with local information via gated sliding-window attention. MELT~\citep{melt2026} shares one KV cache per layer across recursions and updates it via learned gate. Looped Latent Attention~\citep{lla2026} compresses loop-specific keys and values into low-rank codes and reconstructs them when attention reads the cache. The training procedure proposed by \citet{gelberg2026kvcat} masks context-token KV slots during continued pretraining, encouraging the model to learn representations that remain useful when the token-level cache is subsequently compressed by post-hoc methods. These studies address complementary aspects of recurrent computation. Our work studies a fixed write/reuse schedule across iterations of the same weight-shared attention layer.
\paragraph{Recurrent sequence mixers.} Gated DeltaNet~\citep{yang2024gateddeltanet} and mLSTM~\citep{beck2024xlstm} provide recurrent state updates of fixed size, which avoids the sequence-length-dependent KV storage of softmax attention. In the looped hybrid, these mixers do not replace exact attention; instead, they carry computation between Transformer anchors. This permits us to study how often token-level memory must be reconstructed and how much of the effective depth we can implement through recurrent state transitions.

\section{Method}
\label{sec:method}
\paragraph{Looped Hybrid}
\label{sec:looped_hybrid}
The looped hybrid architecture consists of exact-attention Transformer anchors separated by recurrent sequence mixer blocks. The complete logical stack is repeated, and Transformer parameters are shared across recurrence iterations, while recurrent mixer parameters are iteration-specific, and MLP parameters are shared across loops at the same vertical position. This architecture directly exposes the trade-off between exact token-level memory and compressed recurrent computation. Reducing the number of Transformer anchors lowers the number of attention caches, but also lengthens the interval over which recurrent mixers must propagate and transform information without reconstructing exact token-level interactions.

\begin{figure}[t]
    \centering
    \includegraphics[width=0.70\linewidth]{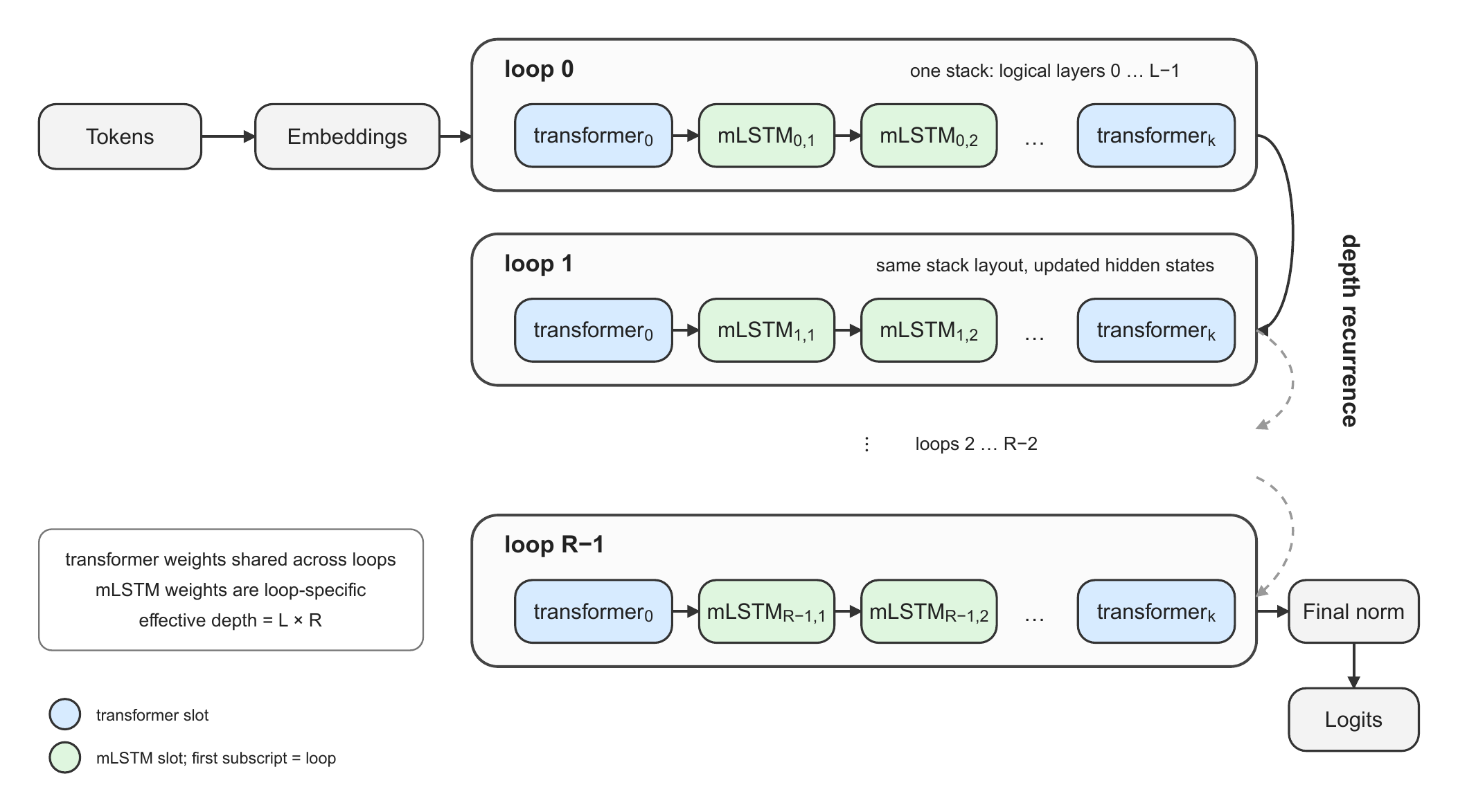}
    \caption{
        Looped hybrid model. Exact-attention Transformer anchors are interleaved with recurrent sequence mixers; the complete logical stack is applied across multiple recurrence iterations.
    }
    \label{fig:looped_hybrid}
\end{figure}

Let the logical stack contain $L$ vertical positions and execute for $R$ recurrence iterations. Denote the hidden state at recurrence iteration $r$ and vertical position $l$ by $h_{r,l}$. A set $\mathcal{A}\subseteq\{0,\ldots,L-1\},$ specifies the positions occupied by exact-attention Transformer anchors. Layers outside $\mathcal{A}$ are implemented using recurrent sequence mixers.
For $l\in\mathcal{A}$, the update is
\[
h_{r,l+1} = T_l(h_{r,l}; \theta_l^T),
\]
where the same parameters $\theta_l^T$ are reused for every $r$. The attention parameters are fixed layer parameters. At recurrent positions $l\notin\mathcal{A}$, the update is
\[
u_{r,l}=\mathcal{M}_{r,l}\left(h_{r,l};\theta^{M}_{r,l}\right),\qquad h_{r,l+1}=u_{r,l}+\mathcal{F}_{l}\left(u_{r,l};\theta^{F}_{l}\right).
\]
where $\mathcal{M}_{r,l}$ is an mLSTM~\citep{beck2024xlstm} or GDN~\citep{yang2024gateddeltanet} mixer and $\mathcal{F}_{l}$ is the MLP assigned to the vertical position $l$. We allocate separate recurrent mixer parameters $\theta^{M}_{r,l}$ for each recurrence iteration, whereas $\theta^{F}_{l}$ is shared across recurrence iterations.
The final hidden state of one logical pass initialises the next recurrence iteration.
A transformer-anchor stride $s_d$ places Transformer anchors every $s_d$ layers; the final layer can optionally also be made an anchor. Decreasing $s_d$ increases the frequency with which exact token-level interactions are reconstructed, but also increases attention computation and the number of KV-producing positions. Increasing $s_d$ delegates longer spans of the stack to fixed-size recurrent mixers.

\paragraph{\textsc{Parcae} and \textsc{Parcae} Wrapper Modes}
\label{sec:parcae}
Unlike the looped hybrid, which repeats an entire logical stack, \textsc{Parcae} confines recurrence to a dedicated latent workspace between fixed Transformer stages.
Our implementation follows the overall \textsc{Parcae} design~\citep{prairie2026parcae}, including the prelude/core/coda decomposition, recurrent workspace, input-injection adapter, stochastic recurrence schedule, and truncated-backpropagation training. Unlike the original architecture, we instantiate the core with mLSTM-based modules~\citep{beck2024xlstm}. This allows a direct comparison between the \textsc{Parcae} organisation and the looped-hybrid architecture. Following \textsc{Parcae}~\citep{prairie2026parcae}, we divide computation into a Transformer prelude, a recurrent core, and a Transformer coda. The prelude produces a representation $e$, which is projected into a recurrent workspace.
Each recurrence iteration first applies an input-injection adapter and then an mLSTM recurrent core~\citep{beck2024xlstm}:
\[
\widetilde{z}_{t}=\mathcal{I}\left(z_t,e\right),\qquad
z_{t+1}=\mathcal{C}\left(\widetilde{z}_{t};\theta^{C}\right).
\]
After $R$ recurrence iterations, the recurrent state is projected back to the Transformer hidden dimension and processed once by the Transformer coda. For each training instance, \textsc{Parcae} samples two depths: a prefix depth $n$, used only to advance the recurrent state, and a suffix depth $k$, retained in the autograd graph. The state is first advanced without constructing an autograd graph; optimisation then differentiates through only the final $k$ iterations.
Evaluation uses deterministic recurrence depth.

The \textsc{Parcae} adapter~\citep{prairie2026parcae} is designed to prevent recurrent drift by repeatedly re-injecting the original input representation $e$. Let $x_t$ be the current recurrent state and $e$ be the representation that entered recurrence. The diagonal adapter computes
\[
\mathrm{dt}=\mathrm{softplus}(\mathrm{dt}_{\mathrm{bias}}),\;
A=\exp(A_{\log}),\;
\mathrm{decay}=\exp(-\mathrm{dt}\cdot A),\;
x_{t+1}=x_t\odot\mathrm{decay}+\mathrm{dt}(eB^\top).
\]
At each recurrence iteration, a learned decay retains part of the current state, while a learned projection of the original representation $e$ is injected to counteract recurrent drift.

The wrapper ablation applies the \textsc{Parcae} adapter before each looped-hybrid recurrence pass:
\[
\widetilde{h}_r=\mathcal{I}(h_r,h_0),
\qquad h_{r+1}=\mathcal{H}(\widetilde{h}_r).
\]
Here, \(h_0\) is the representation entering depth recurrence, \(\mathcal{I}\) is the input-injection adapter, and \(\mathcal{H}\) denotes one pass through the looped-hybrid stack. Figure~\ref{fig:parcae_wrapper_appendix} in Appendix~\ref{app:architecture_figures} illustrates this wrapper. The wrapper isolates input injection from the remainder of the \textsc{Parcae} architecture. It does not introduce the prelude/core/coda decomposition, separate recurrent workspace, stochastic recurrence depth, nor truncated backpropagation.
\begin{figure}[t]
    \centering
    \includegraphics[width=0.80\linewidth]{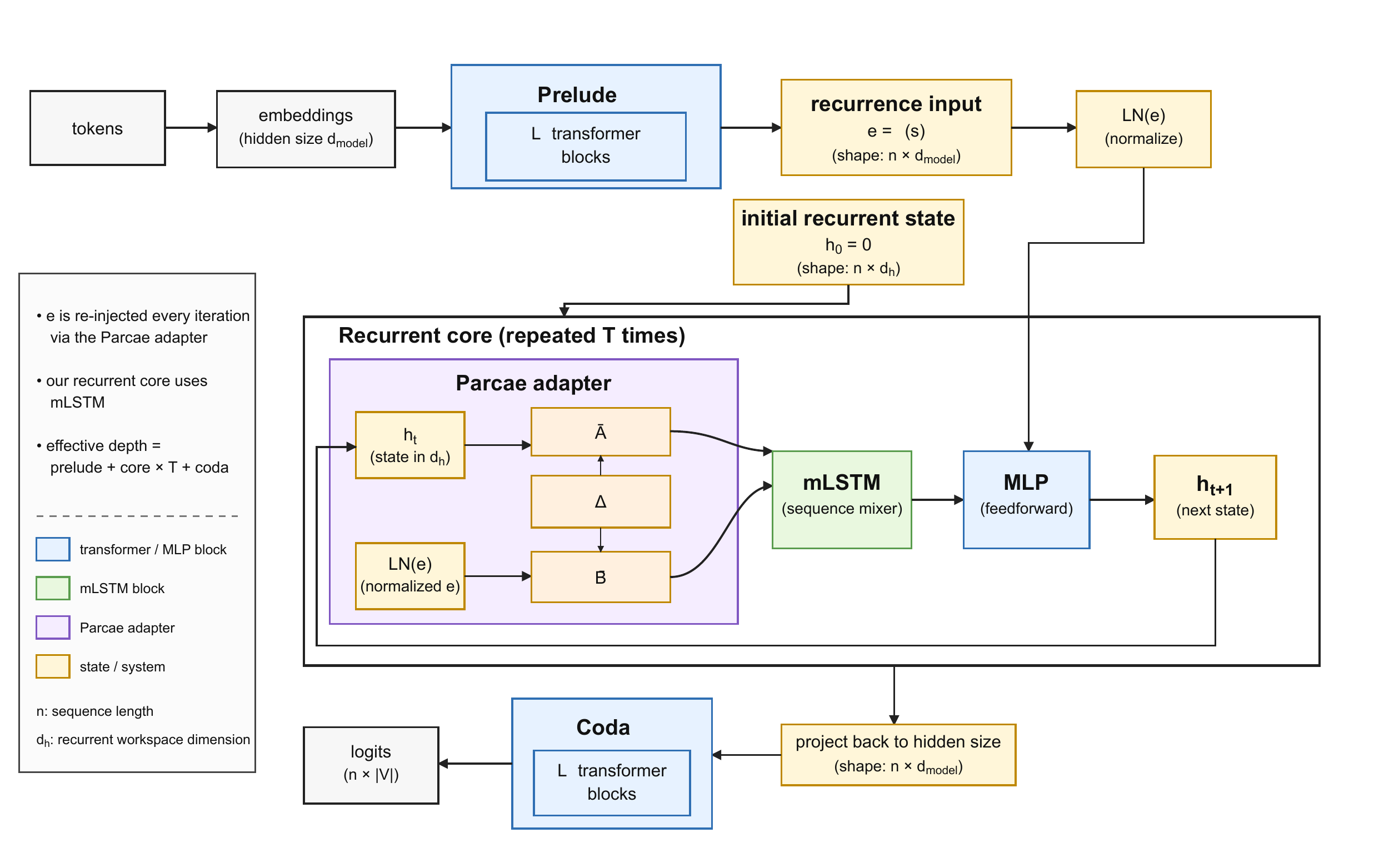}
    \caption{
        In our implementation of the \textsc{Parcae} architecture, Transformer layers encode the input, the representation is projected into a recurrent workspace, a shared mLSTM core is applied repeatedly with stabilising input injection, and the resulting state is projected back before the coda.
    }
    \label{fig:parcae}
\end{figure}
\paragraph{Cache Compression}
\label{sec:cache_compression}
We distinguish two cache-compression regimes. Under \emph{inference-only compression}, the model is trained with ordinary attention~\citep{vaswani2017attention}:
\begin{equation}
\operatorname{Attn}_{r,l}
=
\operatorname{softmax}\left(
\frac{Q_{r,l}K_{\widehat r,l}^{\top}}{\sqrt{d_k}}
\right)V_{\widehat r,l},
\qquad
\widehat r=r \text{ for standard attention},\quad
\widehat r\neq r \text{ for KV replay}.
\label{eq:attention_replay}
\end{equation}
During standard training, keys and values are temporary activations and no persistent cache object is used. When decoding with standard attention, each iteration uses its own $K_{r,l},V_{r,l}$. However, if during decoding a compressed cache policy is applied, the query is computed from the current hidden state but $K_{\widehat r,l},V_{\widehat r,l}$ are reused from an earlier writer iteration. Thus, inference-only compression exposes the model to a different KV provenance at evaluation time.

\emph{Cache-compression training} incorporates the intended sharing policy into the training forward pass. Recurrence iterations are divided into \emph{writers}, which compute and store fresh keys and values, and \emph{readers}, which compute fresh queries but reuse the keys and values of the most recent writer. With cache stride $s_c$, iterations divisible by $s_c$ are writers and use $\widehat r=r$ in Equation~\ref{eq:attention_replay}; all other iterations are readers and use the corresponding earlier writer $\widehat r$. The replay dictionary is not a generation cache. It exists only within one training forward pass and does not contain history from preceding autoregressive decode steps. Its purpose is to reproduce the cross-iteration provenance of keys and values that the model will encounter under compressed decoding.

\begin{figure}[!htbp]
    \centering
    \includegraphics[width=0.80\linewidth]{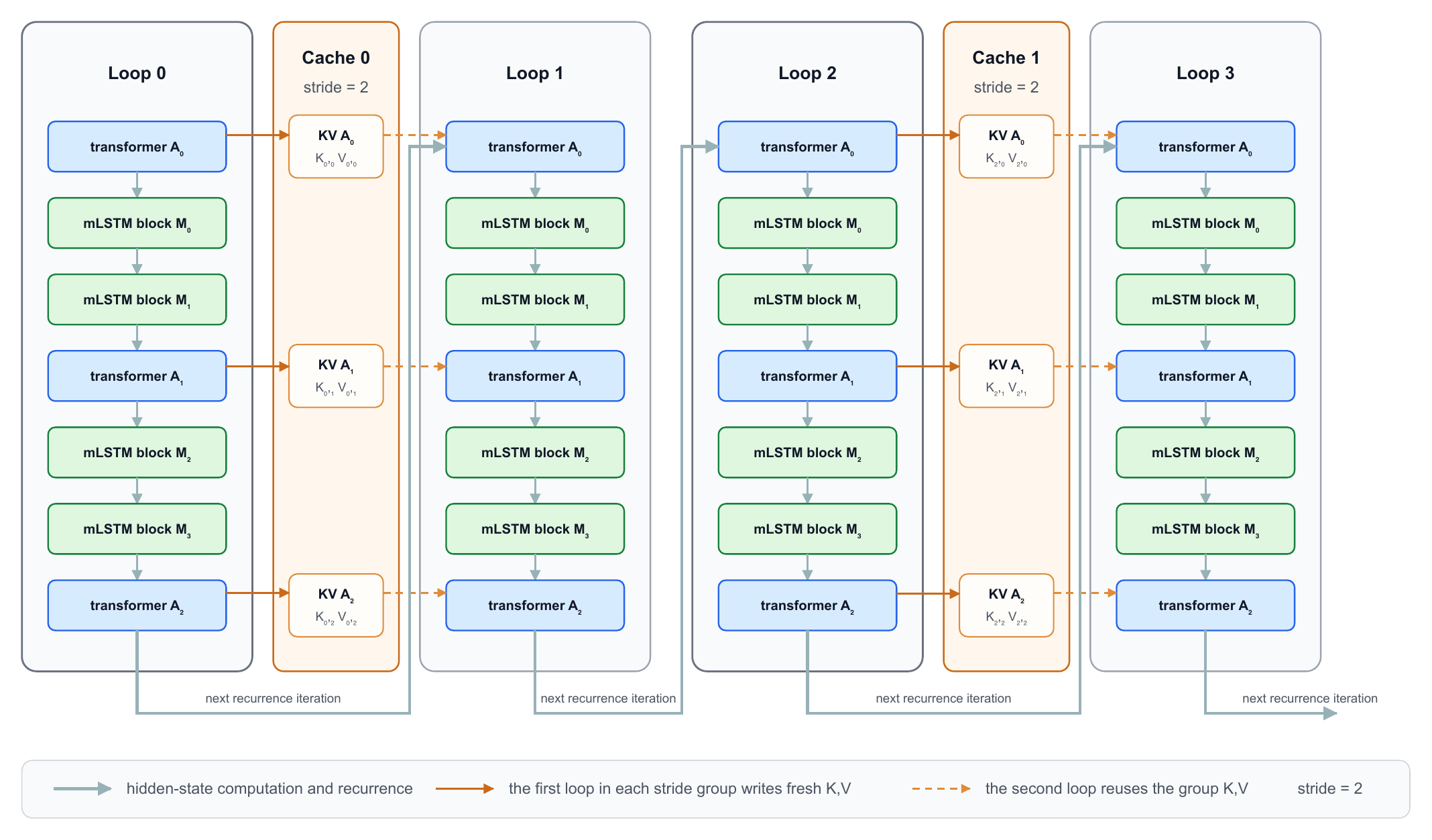}
    \caption{Stride-2 KV-cache routing. Orange connections indicate cache writes, dashed connections indicate cache reuse, and grey connections show hidden-state propagation.}
    \label{fig:cache_stride2}
\end{figure}

\begin{figure}[t]
    \centering
    \includegraphics[width=0.80\linewidth]{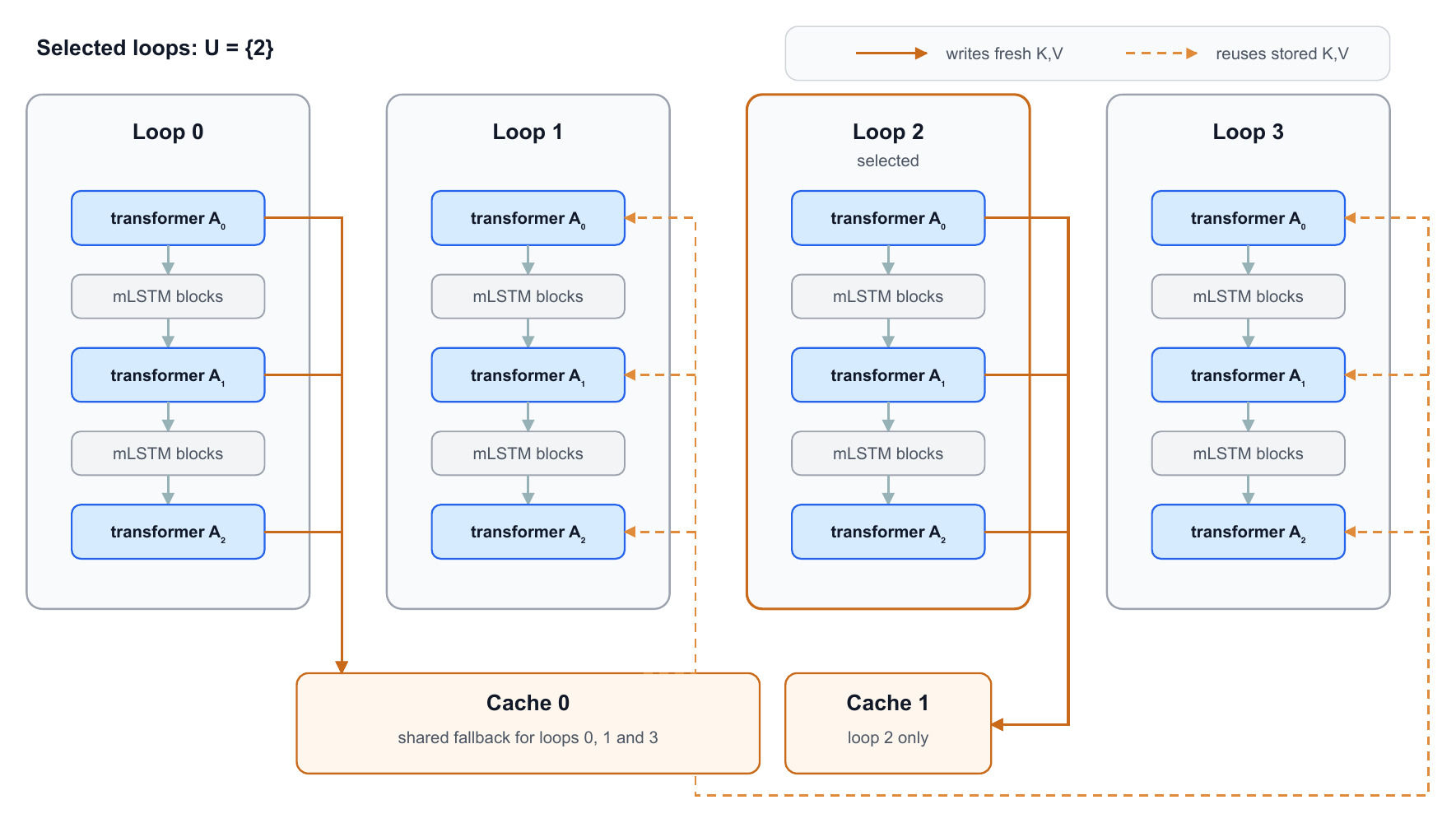}
    \caption{Selective-loop KV-cache routing. Orange connections indicate cache writes, dashed connections indicate cache reuse, and grey connections show hidden-state propagation.}
    \label{fig:cache_selective_loop}
\end{figure}

\paragraph{Cache-Routing Policies}
\label{sec:cache_routing}
Let \(r\in\{0,\ldots,R-1\}\) denote the recurrence iteration, let \(l\in\mathcal{A}\) denote an exact-attention layer, and let \(L\) be the number of logical stack positions. Each policy maps \((r,l)\) to a cache group \(g(r,l)\) and therefore to the flat cache slot \(c(r,l)=g(r,l)L+l\). The write rule \(w(r,l)\in\{0,1\}\) specifies whether iteration \(r\) creates fresh keys and values at layer \(l\) or reuses the tensors already stored in the selected group. Figures~\ref{fig:cache_stride2} and~\ref{fig:cache_selective_loop} illustrate two of the routing policies used in our experiments. Figures of other routing policies are provided in Appendix~\ref{app:cache_routing_figures}.

\textbf{Stride-$s$ cache.} Recurrence iterations are grouped into windows of size $s$, with \(g(r,l)=\lfloor r/s\rfloor\), \(c(r,l)=\lfloor r/s\rfloor L+l\), and \(w(r,l)=\mathbf{1}[r\bmod s=0]\). Only the first recurrence iteration in each stride group writes new keys and values. The shared-cache and per-iteration-cache cases are the special cases \(s=R\) and \(s=1\), respectively. Figure~\ref{fig:cache_stride2} illustrates the routing pattern for $s=2$.

\textbf{Selective-loop cache.}
Let $U$ be the set of recurrence iterations which are selected to receive independent cache storage. Selected loops are assigned their own group using an offset of $+1$; all other loops use the shared fallback group:
\[
g(r,l)=
\begin{cases}
r+1,&r\in U,\\
0,&r\notin U,
\end{cases}
\qquad
w(r,l)=\mathbf{1}[r\in U]\lor\mathbf{1}[r=0].
\]
The offset prevents a selected loop from colliding with the shared fallback group.

\textbf{Selective-layer cache.}
Let $S$ be the set of transformer-anchor layers selected for independent per-loop KV storage. Selected layers receive a separate cache group for each recurrence iteration; other layers use the shared fallback group~0.

\section{Experiments}
\label{sec:experiments}
We organise the experiments around three questions. First, how frequently must a looped-hybrid model reconstruct exact token-level interactions? Second, does restricting recurrence to a dedicated workspace improve over recurrence through the full hidden representation? Third, can a fully looped Transformer learn to operate under a compressed cross-iteration KV topology? The non-recurrent Transformer is included as a baseline in every quantitative comparison.
Parameter counts differ across the looped-hybrid configurations because recurrent mixer parameters are iteration-specific and reported in Appendix~\ref{app:parameter_counts}; the baseline and full looped Transformer contain \(406.37\)M parameters.
Higher is better for all length-normalised accuracy metrics; lower is better for perplexity and memory. Bold entries denote the best observed result in each table column. For CORE and CORE-Extended, we report the mean \(\pm\) standard deviation across evaluation seeds for the selected checkpoint. Additional training and evaluation details are provided in Appendix~\ref{app:setup}.

\paragraph{Baseline.} Our primary baseline is a 24-layer, non-recurrent Transformer trained with sequence length \(4096\), learning rate \(3\times10^{-3}\), and a token budget of \(5.03\)B tokens. It provides a parameter- and training-budget reference for the looped-hybrid and cache-compression experiments. The per-iteration control establishes the reference performance of a looped Transformer without cross-iteration KV compression.

\paragraph{Looped Hybrid.}
The main configuration contains \(24\) logical layers, executes 4 recurrence iterations, and is trained at sequence length \(4096\). The anchor stride \(s_d\) controls the distance between exact-attention Transformer anchors. We also compare a matched \(12\)-layer, \(8\)-iteration configuration used for the \textsc{Parcae} architecture and evaluate a wrapper ablation that applies the \textsc{Parcae} adapter without the native \textsc{Parcae} architecture.
\paragraph{\textsc{Parcae} Implementation with mLSTM.}
We followed the original \textsc{Parcae} training configuration~\citep{prairie2026parcae} as faithfully as permitted by our infrastructure. In particular, the reference configuration uses 8 recurrence iterations, hidden and recurrent dimensions of \(1024\), backpropagation depth \(4\), gradient clipping at \(1.0\), no warm-up, and a 50\% cool-down. Our principal difference is the training budget: the original \(370\)M-parameter configuration specifies \(29.6\)B tokens; models here are generally trained on \(5.03\)B tokens. We also use our Nemotron-CC-HQ data pipeline, tokeniser, and AdamW optimiser rather than the original FineWeb-Edu, tokeniser, and Muon setup. The comparison should therefore be interpreted as an architectural study under our common budget, not as a reproduction of the published \textsc{Parcae} results. Changing the recurrent workspace dimension \(d_r\) also changes the number of model parameters and the computation performed by each recurrent-core application. The \(d_r\in\{768,1024,1536\}\) configurations contain \(241.10\)M, \(259.76\)M, and \(315.94\)M parameters, respectively; exact counts are reported in Appendix~\ref{app:parameter_counts}. We additionally train the \(2/4/6,\;d_r=1536\) configuration for the original \textsc{Parcae} budget of \(29.6\)B tokens.

\paragraph{KV Replay and Reasoning.}
We evaluate a fully looped Transformer with \(24\) physical layers and 6 recurrence iterations. Models trained with recurrent KV replay are compared with those trained using ordinary attention and compressed only during evaluation. Stride \(2\) retains three recurrence-dependent KV groups, stride \(3\) retains two. We evaluate both matched and mismatched train--inference topologies. Additionally, we evaluate GSM8K~\citep{cobbe2021training} using 8-shot chain-of-thought prompting with 8 generations per problem and report the mean Majority@8 and Pass@8 over all problems together with 95\% percentile bootstrap confidence intervals from 10\,000 resamples of the GSM8K evaluation set.

\section{Results and Discussion}
\label{sec:results}

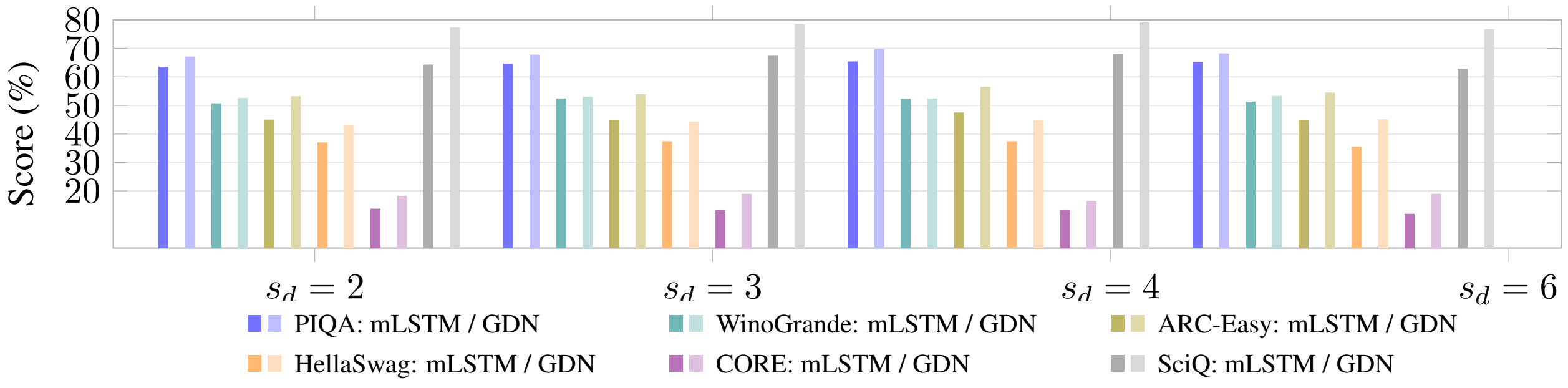
\begin{figure*}[t]
\centering
\begin{tikzpicture}
\begin{axis}[
    ybar,
    bar width=2.6pt,
    width=\textwidth,
    height=4.0cm,
    ymin=0,
    ymax=80,
    ytick={20,30,40,50,60,70,80},
    ylabel={Score (\%)},
    xtick={0,0.75,1.5,2.25},
    xticklabels={$s_d=2$,$s_d=3$,$s_d=4$,$s_d=6$},
    xmin=-0.38,
    xmax=2.35,
    enlarge x limits=false,
    ymajorgrids=true,
    grid style={gray!20},
    axis line style={gray!60},
    tick style={gray!60},
    legend style={
        at={(0.5,-0.23)},
        anchor=north,
        draw=none,
        font=\scriptsize,
        legend columns=3,
        column sep=9pt,
        row sep=1pt,
        cells={anchor=west}
    }
]

\addplot+[fill=blue!55,draw=blue!55,forget plot]
coordinates {(-0.300,63.3) (0.350,64.4) (1.000,65.2) (1.650,64.9)};
\addplot+[fill=blue!25,draw=blue!25,forget plot]
coordinates {(-0.250,66.9) (0.400,67.6) (1.050,69.6) (1.700,68.0)};

\addplot+[fill=teal!55,draw=teal!55,forget plot]
coordinates {(-0.200,50.5) (0.450,52.2) (1.100,52.1) (1.750,51.1)};
\addplot+[fill=teal!25,draw=teal!25,forget plot]
coordinates {(-0.150,52.4) (0.500,52.8) (1.150,52.2) (1.800,53.1)};

\addplot+[fill=olive!60,draw=olive!60,forget plot]
coordinates {(-0.100,44.8) (0.550,44.7) (1.200,47.3) (1.850,44.7)};
\addplot+[fill=olive!30,draw=olive!30,forget plot]
coordinates {(-0.050,53.0) (0.600,53.7) (1.250,56.3) (1.900,54.3)};

\addplot+[fill=orange!55,draw=orange!55,forget plot]
coordinates {(0.000,36.8) (0.650,37.2) (1.300,37.2) (1.950,35.3)};
\addplot+[fill=orange!25,draw=orange!25,forget plot]
coordinates {(0.050,43.0) (0.700,44.1) (1.350,44.6) (2.000,44.9)};

\addplot+[fill=violet!55,draw=violet!55,forget plot]
coordinates {(0.100,13.560) (0.750,13.080) (1.400,13.160) (2.050,11.760)};
\addplot+[fill=violet!25,draw=violet!25,forget plot]
coordinates {(0.150,18.080) (0.800,18.790) (1.450,16.260) (2.100,18.790)};

\addplot+[fill=gray!65,draw=gray!65,forget plot]
coordinates {(0.200,64.1) (0.850,67.4) (1.500,67.7) (2.150,62.6)};
\addplot+[fill=gray!30,draw=gray!30,forget plot]
coordinates {(0.250,77.1) (0.900,78.2) (1.550,78.9) (2.200,76.5)};

\addlegendimage{legend image code/.code={
    \path[use as bounding box] (0,0) rectangle (0,0);
}}
\addlegendentry{%
    \tikz[baseline=-0.6ex]
    \fill[blue!55] (0,-2pt) rectangle (4pt,3pt);
    \tikz[baseline=-0.6ex]
    \fill[blue!25] (0,-2pt) rectangle (4pt,3pt);
    \hspace{2pt}PIQA: mLSTM / GDN%
}

\addlegendimage{legend image code/.code={
    \path[use as bounding box] (0,0) rectangle (0,0);
}}
\addlegendentry{%
    \tikz[baseline=-0.6ex]
    \fill[teal!55] (0,-2pt) rectangle (4pt,3pt);
    \tikz[baseline=-0.6ex]
    \fill[teal!25] (0,-2pt) rectangle (4pt,3pt);
    \hspace{2pt}WinoGrande: mLSTM / GDN%
}

\addlegendimage{legend image code/.code={
    \path[use as bounding box] (0,0) rectangle (0,0);
}}
\addlegendentry{%
    \tikz[baseline=-0.6ex]
    \fill[olive!60] (0,-2pt) rectangle (4pt,3pt);
    \tikz[baseline=-0.6ex]
    \fill[olive!30] (0,-2pt) rectangle (4pt,3pt);
    \hspace{2pt}ARC-Easy: mLSTM / GDN%
}

\addlegendimage{legend image code/.code={
    \path[use as bounding box] (0,0) rectangle (0,0);
}}
\addlegendentry{%
    \tikz[baseline=-0.6ex]
    \fill[orange!55] (0,-2pt) rectangle (4pt,3pt);
    \tikz[baseline=-0.6ex]
    \fill[orange!25] (0,-2pt) rectangle (4pt,3pt);
    \hspace{2pt}HellaSwag: mLSTM / GDN%
}

\addlegendimage{legend image code/.code={
    \path[use as bounding box] (0,0) rectangle (0,0);
}}
\addlegendentry{%
    \tikz[baseline=-0.6ex]
    \fill[violet!55] (0,-2pt) rectangle (4pt,3pt);
    \tikz[baseline=-0.6ex]
    \fill[violet!25] (0,-2pt) rectangle (4pt,3pt);
    \hspace{2pt}CORE: mLSTM / GDN%
}

\addlegendimage{legend image code/.code={
    \path[use as bounding box] (0,0) rectangle (0,0);
}}
\addlegendentry{%
    \tikz[baseline=-0.6ex]
    \fill[gray!65] (0,-2pt) rectangle (4pt,3pt);
    \tikz[baseline=-0.6ex]
    \fill[gray!30] (0,-2pt) rectangle (4pt,3pt);
    \hspace{2pt}SciQ: mLSTM / GDN%
}

\end{axis}
\end{tikzpicture}

\caption{Looped-hybrid results on the main setup: we report normalised accuracies on PIQA, WinoGrande, ARC-Easy, HellaSwag, and SciQ, together with CORE; \(s_d\) denotes the Transformer-anchor stride.}
\label{fig:looped_hybrid_results}
\end{figure*}

\paragraph{Looped Hybrid with mLSTM and Gated DeltaNet Mixers}
\label{sec:results_looped_hybrid}
Figure~\ref{fig:looped_hybrid_results} examines the effect of recurrent mixer choice in the looped-hybrid architecture and its sensitivity to the spacing of the remaining attention layers. Full results, including perplexity and the wrapper ablation, are provided in Appendix~\ref{app:additional_results}, Table~\ref{tab:full_results}.

For mLSTM hybrids, models with denser Transformer anchors achieve higher downstream accuracy and lower perplexity than those with larger anchor strides, but the results do not establish a monotonic relationship among the strides.
Applying the \textsc{Parcae} adapter as an isolated wrapper degrades accuracy. The input-reinjection wrapper alone is therefore insufficient to improve the looped-hybrid architecture under the evaluated training setting. In our settings, the looped-hybrid performance depends strongly on the choice of recurrent mixer, and replacing the mLSTM mixers with Gated DeltaNet substantially improves the results.

\paragraph{\textsc{Parcae} Implementation with mLSTM}
\label{sec:results_parcae}
Table~\ref{tab:parcae_results} compares two prelude/core/coda allocations and recurrent workspace dimensions \(d_r\in\{768,1536\}\); the complete sweep, including \(d_r=1024\) and the \(6/4/2\) allocation, is provided in Appendix~\ref{app:parcae_results}. Increasing the recurrent workspace dimension does not produce a uniform improvement across the two allocations. The \(29.6\)B-token run improves every reported metric relative to the same \(2/4/6,\;d_r=1536\) configuration trained for \(5.03\)B tokens.

\begin{table*}[t]
    \centering
    \caption{Selected results of our \textsc{Parcae} implementation. Architecture denotes the prelude/core/coda allocation, \(d_r\) is the recurrent workspace dimension. The non-recurrent Transformer uses the main \(24\)-layer setup. CORE metrics are reported in percentage points.}
    \label{tab:parcae_results}
    \scriptsize
    \setlength{\tabcolsep}{2.7pt}
    \resizebox{\textwidth}{!}{%
    \begin{tabular}{lccccccccccc}
        \toprule
        Model
        & \(d_r\)
        & Arch.
        & PIQA-N
        & Wino.
        & ARC-E-N
        & Hella.-N
        & ARC-C-N
        & SciQ-N
        & CORE
        & CORE Ext.
        & WikiText PPL \\
        \midrule

        Transformer baseline
        & --
        & --
        & \textbf{0.664}
        & 0.500
        & \textbf{0.497}
        & \textbf{0.369}
        & \textbf{0.276}
        & \textbf{0.713}
        & \(\mathbf{14.95\pm0.29}\)
        & \(\mathbf{10.26\pm0.39}\)
        & \textbf{35.988} \\

        \midrule

        \textsc{Parcae} with mLSTM
        & 768
        & \(2/4/6\)
        & 0.634
        & 0.502
        & 0.440
        & 0.313
        & 0.256
        & 0.642
        & \(10.50\pm0.17\)
        & \(7.57\pm0.13\)
        & 48.741 \\

        \textsc{Parcae} with mLSTM
        & 1536
        & \(2/4/6\)
        & 0.632
        & 0.510
        & 0.436
        & 0.323
        & 0.236
        & 0.668
        & \(12.47\pm0.16\)
        & \(8.80\pm0.39\)
        & 45.289 \\

        \midrule

        \textsc{Parcae} with mLSTM
        & 768
        & \(4/4/4\)
        & 0.632
        & \textbf{0.524}
        & 0.439
        & 0.313
        & 0.266
        & 0.631
        & \(12.06\pm0.17\)
        & \(8.90\pm0.23\)
        & 48.700 \\

        \textsc{Parcae} with mLSTM
        & 1536
        & \(4/4/4\)
        & 0.633
        & 0.515
        & 0.449
        & 0.326
        & 0.247
        & 0.650
        & \(12.09\pm0.15\)
        & \(8.42\pm0.11\)
        & 45.460 \\

        \midrule
        \multicolumn{12}{c}{Extended training (original \textsc{Parcae} budget)}\\
        \midrule

        \textsc{Parcae} with mLSTM
        & 1536
        & \(2/4/6\)
        & 0.654
        & 0.513
        & 0.492
        & 0.369
        & 0.276
        & 0.719
        & \(15.10\pm0.06\)
        & \(9.52\pm0.25\)
        & 35.750 \\

        \bottomrule
    \end{tabular}%
    }
\end{table*}

\paragraph{Cache-Compression Training}
\label{sec:results_cache_training}

\begin{table*}[t]
    \centering
    \caption{
        Cross-iteration KV compression in a 6-iteration looped
        Transformer. The evaluation column specifies the decoding cache stride. CORE metrics are reported in percentage points.
    }
    \label{tab:cache_training_results}
    \scriptsize
    \setlength{\tabcolsep}{2.7pt}
    \resizebox{\textwidth}{!}{%
    \begin{tabular}{llcccccccccc}
        \toprule
        Training
        & Eval.
        & PIQA-N
        & Wino.
        & ARC-E-N
        & Hella.-N
        & ARC-C-N
        & SciQ-N
        & CORE
        & CORE Ext.
        & WikiText PPL \\
        \midrule

        Transformer baseline
        & --
        & 0.664
        & 0.500
        & 0.497
        & 0.369
        & 0.276
        & 0.713
        & \(14.95\pm0.29\)
        & \(10.26\pm0.39\)
        & 35.988 \\

        Full looped Transformer
        & Per-iteration
        & \textbf{0.678}
        & 0.528
        & \textbf{0.542}
        & \textbf{0.429}
        & 0.303
        & 0.782
        & \(\mathbf{18.96\pm0.23}\)
        & \(\mathbf{12.58\pm0.09}\)
        & 31.204 \\

        \midrule

        Replay \(s=2\)
        & \(s=2\)
        & 0.660
        & 0.502
        & 0.508
        & 0.393
        & 0.283
        & 0.739
        & \(16.76\pm0.07\)
        & \(11.07\pm0.29\)
        & 34.214 \\

        Replay \(s=2\)
        & \(s=3\)
        & 0.606
        & 0.510
        & 0.374
        & 0.342
        & 0.279
        & 0.548
        & \(11.29\pm0.02\)
        & \(8.59\pm0.05\)
        & 162.992 \\

        Replay \(s=3\)
        & \(s=2\)
        & 0.670
        & 0.510
        & 0.512
        & 0.424
        & \textbf{0.310}
        & 0.698
        & \(11.34\pm0.15\)
        & \(8.24\pm0.05\)
        & 60.969 \\

        Replay \(s=3\)
        & \(s=3\)
        & \textbf{0.678}
        & \textbf{0.542}
        & 0.532
        & 0.418
        & 0.296
        & \textbf{0.784}
        & \(18.71\pm0.16\)
        & \(12.41\pm0.16\)
        & \textbf{31.095} \\

        \midrule

        Standard
        & \(s=2\)
        & 0.651
        & 0.495
        & 0.460
        & 0.385
        & 0.282
        & 0.611
        & \(14.48\pm0.10\)
        & \(10.39\pm0.18\)
        & 83.867 \\

        Standard
        & \(s=3\)
        & 0.615
        & 0.496
        & 0.388
        & 0.347
        & 0.285
        & 0.483
        & \(11.66\pm0.09\)
        & \(8.86\pm0.14\)
        & 180.065 \\

        \bottomrule
    \end{tabular}%
    }
\end{table*}
Table~\ref{tab:cache_training_results} compares replay-trained models with models compressed only during evaluation. For both cache strides, matched replay training outperforms inference-only cache compression when training and evaluation use the same cache topology. A checkpoint can be evaluated with different replay strides, but our results show that this does not preserve performance
equally across topologies.
\begin{nxaiinfo}
The strongest observed results are obtained by replay training with stride \(3\), despite retaining fewer KV groups than stride \(2\). For the evaluated configurations, more aggressive sharing is therefore not inherently destructive when it is represented during optimisation.
\end{nxaiinfo}

\begin{table*}[t]
    \centering
    \begin{minipage}[t]{0.48\textwidth}
        \centering
        \caption{GSM8K reasoning with cross-iteration KV sharing. Values are point estimates with 95\% percentile bootstrap confidence intervals.}
        \label{tab:gsm8k_results}
        \scriptsize
        \setlength{\tabcolsep}{2pt}
        \begin{tabular}{lccc}
            \toprule
            Training & Eval. & Majority@8 & Pass@8 \\
            \midrule
            Replay \(s=2\) & \(s=2\) & 0.016 [0.010, 0.023] & 0.080 [0.066, 0.096] \\
            Replay \(s=2\) & \(s=3\) & 0.005 [0.002, 0.010] & 0.039 [0.029, 0.049] \\
            Replay \(s=3\) & \(s=3\) & \textbf{0.024 [0.017, 0.033]} & \textbf{0.092 [0.077, 0.108]} \\
            Standard & \(s=2\) & 0.017 [0.010, 0.024] & 0.072 [0.058, 0.086] \\
            Standard & \(s=3\) & 0.006 [0.002, 0.011] & 0.046 [0.034, 0.057] \\
            \bottomrule
        \end{tabular}
            \end{minipage}
    \hfill
    \begin{minipage}[t]{0.48\textwidth}
        \centering
        \caption{Peak evaluation-memory delta during 8-shot GSM8K decoding. Stride \(2\) retains three recurrence-dependent KV groups; stride \(3\) retains two.}
        \label{tab:memory_results}
        \scriptsize
        \setlength{\tabcolsep}{3pt}
        \begin{tabular}{lc}
            \toprule
            Model and cache topology & Peak memory delta (MB) \\
            \midrule
            Transformer baseline & \(\mathbf{2205.1}\) \\
            \midrule
            Looped Transformer, stride \(2\) & \(3241.5\pm350.7\) \\
            Looped Transformer, stride \(3\) & \(2922.9\pm348.3\) \\
            \bottomrule
        \end{tabular}
    \end{minipage}
\end{table*}

Table~\ref{tab:gsm8k_results} reports Majority@8 and Pass@8 together with 95\% bootstrap confidence intervals computed from 10\,000 resamples of the GSM8K evaluation set. Matched replay training improves reasoning under compressed decoding, with the strongest results obtained when both training and evaluation use stride~3.
\begin{nxaiwarning}[Caution]
However, the absolute GSM8K scores are low, and each configuration uses a single training seed. These results should therefore be treated as a comparison among cache policies and they do not establish that the observed ordering is robust across training runs.
\end{nxaiwarning}

\paragraph{Memory Analysis}
\label{sec:results_memory}

Table~\ref{tab:memory_results} reports the maximum GPU memory usage observed during eight-shot GSM8K decoding, measured relative to GPU memory usage immediately before launching the decoding process. Values for the looped Transformer are aggregated over evaluation seeds \(42\), \(43\), and \(44\). The full per-iteration-cache configuration with 6 iterations was computationally unfeasible in our setup. Within the recurrent models, reducing the number of KV groups lowers the measured peak but the difference is smaller than either standard deviation. The measurement also includes parameters, temporary activations, allocator overhead, and other states unaffected by cache routing.

\noindent
\begin{minipage}[c]{0.61\columnwidth}
    \small
    The analytical KV-cache sizes in bytes are computed as
    \[
    \underbrace{2}_{K\text{ and }V}
    \times
    \underbrace{24}_{\text{layers}}
    \times
    \underbrace{G}_{\text{KV groups}}
    \times
    \underbrace{16}_{\text{KV heads}}
    \times
    \underbrace{64}_{\text{head dimension}}
    \times
    \underbrace{2}_{\text{bytes per BF16 element}}.
    \]
\end{minipage}
\hfill
\begin{minipage}[c]{0.35\columnwidth}
    \centering
    \scriptsize
    \setlength{\tabcolsep}{4pt}
    \begin{tabular}{lcc}
        \toprule
        Cache policy & \(G\) & KV (KiB/token) \\
        \midrule
        Per iteration & 6 & 576 \\
        Stride 2 & 3 & 288 \\
        Stride 3 & 2 & 192 \\
        \bottomrule
    \end{tabular}
\end{minipage}

\section{Conclusion}
Depth recurrence reduces parameter growth but introduces additional inference memory through the KV cache. We studied three approaches to reducing this memory cost: replacing most attention layers with recurrent mixers, restricting recurrence to a dedicated latent workspace, and training a looped Transformer to reuse KV memory across recurrence iterations. In the looped-hybrid architecture, the effect of Transformer-anchor spacing depends on the recurrent mixer. For mLSTM, denser Transformer anchors are associated with higher downstream accuracy and lower language-modelling perplexity in the evaluated configurations, although the results do not establish a monotonic relationship across strides. Gated DeltaNet is comparatively robust across the evaluated spacings. Across the evaluated \textsc{Parcae} configurations, no single architecture dominates all metrics, and the effect of recurrent workspace dimension depends on the allocation of Transformer layers around the recurrent core. The strongest observed result comes from recurrent KV replay. Replay-trained models outperform inference-only cache compression when training and evaluation use the same cache topology, with the strongest overall results obtained for the evaluated stride-3 configuration. Thus in our setting, cross-iteration KV sharing can be incorporated into training while retaining performance close to the full looped Transformer with separate KV caches.

\paragraph{Discussion and Future Work.}
\label{sec:limitations}
The baseline and full looped Transformer contain 406.37M parameters, while the looped-hybrid configurations range from 965M to 1.47B parameters. 
Additionally, our experiments compare complete configurations rather than isolating every architectural factor. The results are therefore exploratory rather than resource-matched comparisons. The relative ordering of architectures and policies may change at larger scales, with different datasets, or under longer training. The replay results are evaluated only for fixed deterministic cache-sharing schedules and do not establish that the observed improvements extend to token-dependent routing or policies in which different layers use different training-time replay schedules. We report GSM8K performance only as a controlled comparison of cache policies; it should not be interpreted as evidence of strong mathematical reasoning. Reported memory measurements used to establish relative comparisons between routing policies are implementation-level peak values obtained within our evaluation pipeline and include runtime overhead beyond the KV cache itself.

Our work is foundational and has no specific deployment target. More memory-efficient depth-recurrent architectures could reduce the computational resources required for downstream language-model applications, and potentially make such applications easier to deploy at scale; their broader effects therefore depend on the downstream systems in which they are used.

\clearpage
\section*{Acknowledgements}
This work was supported by European Union’s Horizon Europe research and innovation programme under grant agreement number 101214398 (ELLIOT). We acknowledge EuroHPC Joint Undertaking for awarding us access to Leonardo at CINECA, Italy, and Deucalion at MACC, Portugal.

\bibliographystyle{plainnat}
\bibliography{references}

\clearpage
\appendix
\startcontents[appendix]

\section*{Appendix}
\printcontents[appendix]{l}{1}{\setcounter{tocdepth}{2}}
\clearpage

\appendix

\section{Additional Figures}

\subsection{\textsc{Parcae} Wrapper}
\label{app:architecture_figures}

\begin{figure}[!htbp]
    \centering
    \includegraphics[width=0.90\linewidth]{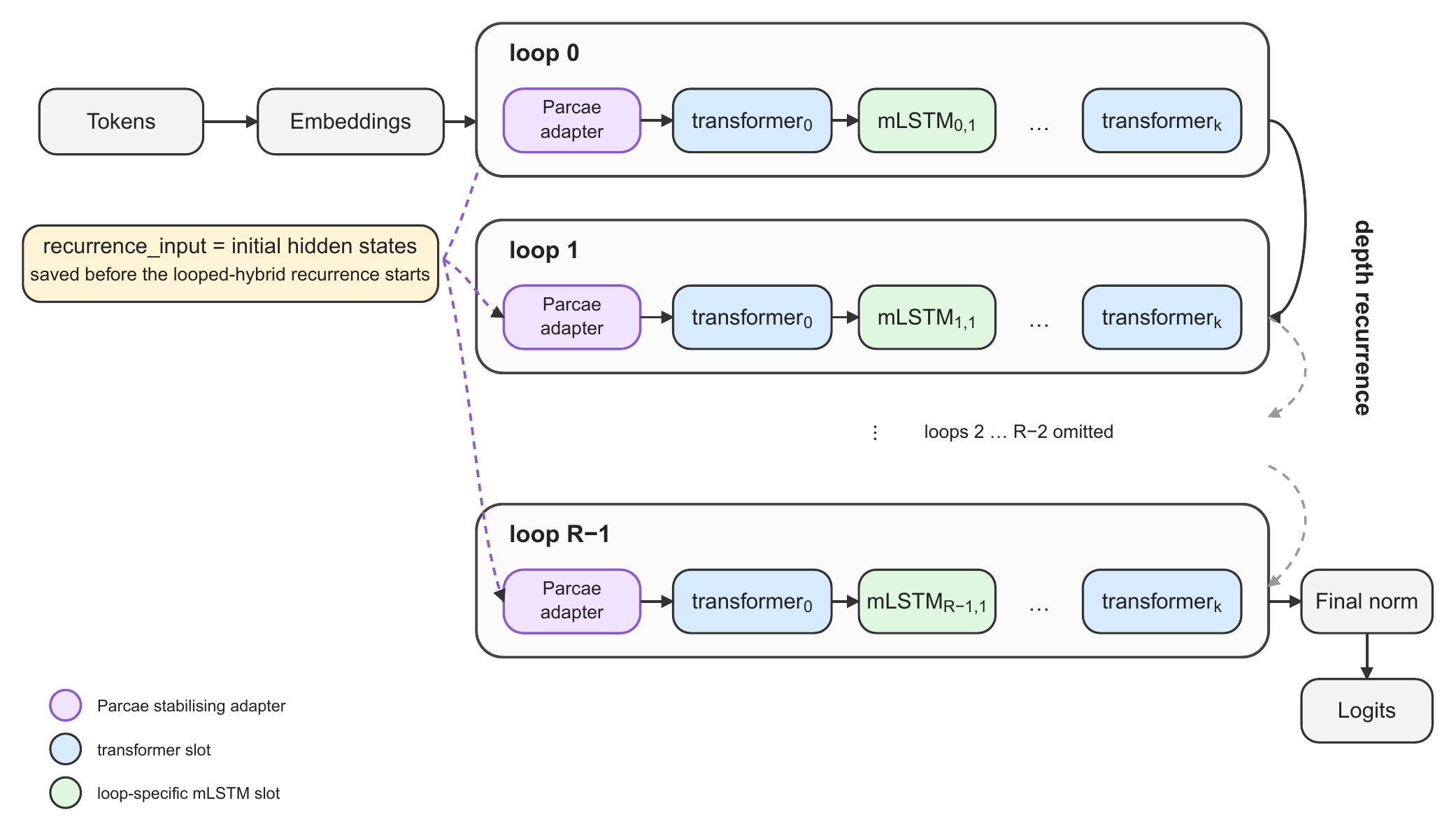}
    \caption{
        \textsc{Parcae}-style adapter applied as a wrapper around the Looped Hybrid recurrence. The adapter repeatedly injects the representation entering recurrence but does not introduce the \textsc{Parcae} prelude/core/coda decomposition.
    }
    \label{fig:parcae_wrapper_appendix}
\end{figure}

Figure~\ref{fig:parcae_wrapper_appendix} illustrates the wrapper ablation described in Section~\ref{sec:parcae}.

\subsection{Additional Cache-Routing Policies}
\label{app:cache_routing_figures}

\begin{figure}[!htbp]
    \centering
    \includegraphics[width=0.80\linewidth]{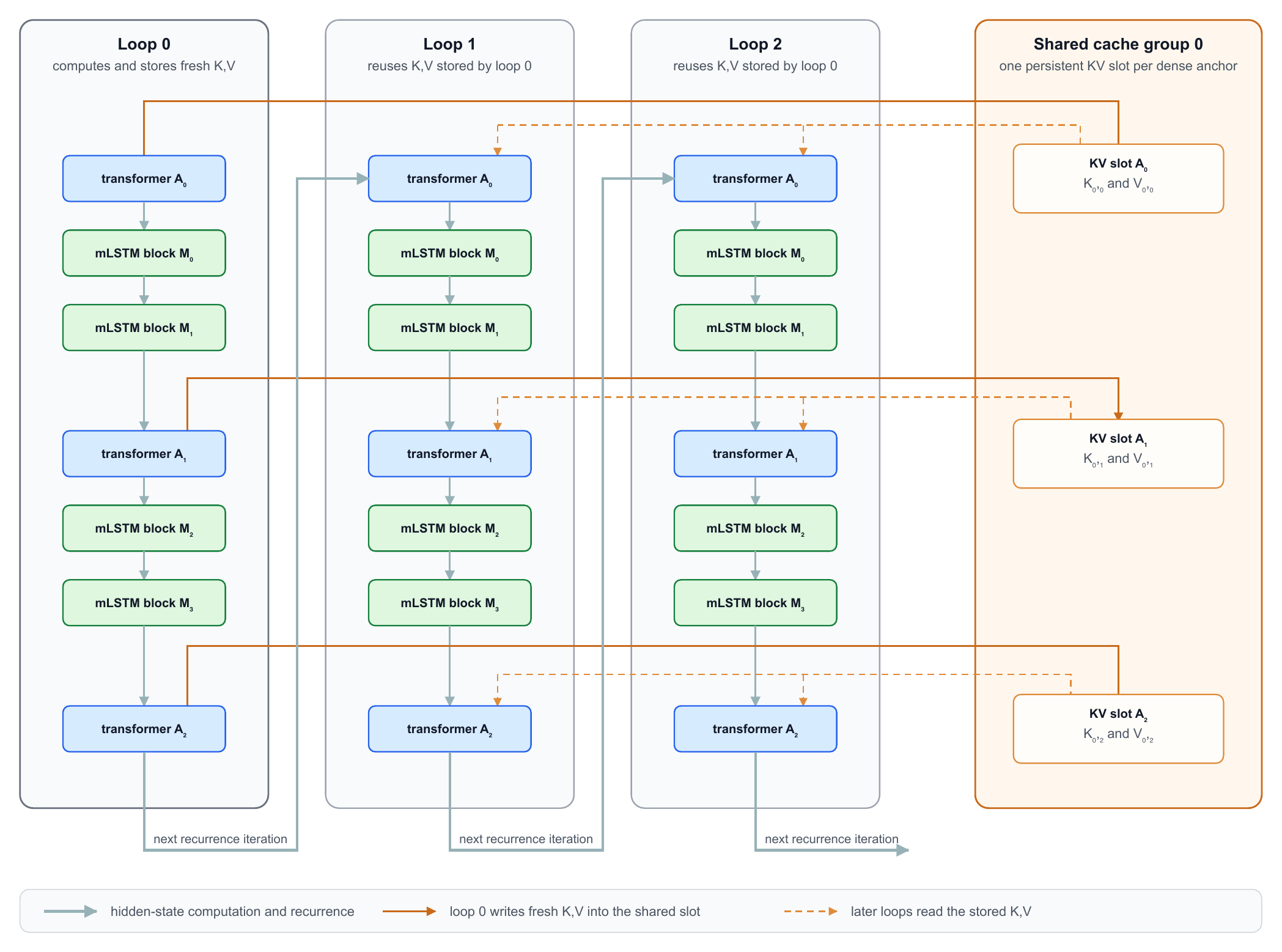}
    \caption{
        Shared KV-cache routing. Only the first recurrence iteration writes keys and values; all subsequent iterations reuse the same cache group.
    }
    \label{fig:cache_shared_appendix}
\end{figure}

\begin{figure}[!htbp]
    \centering
    \includegraphics[width=0.90\linewidth]{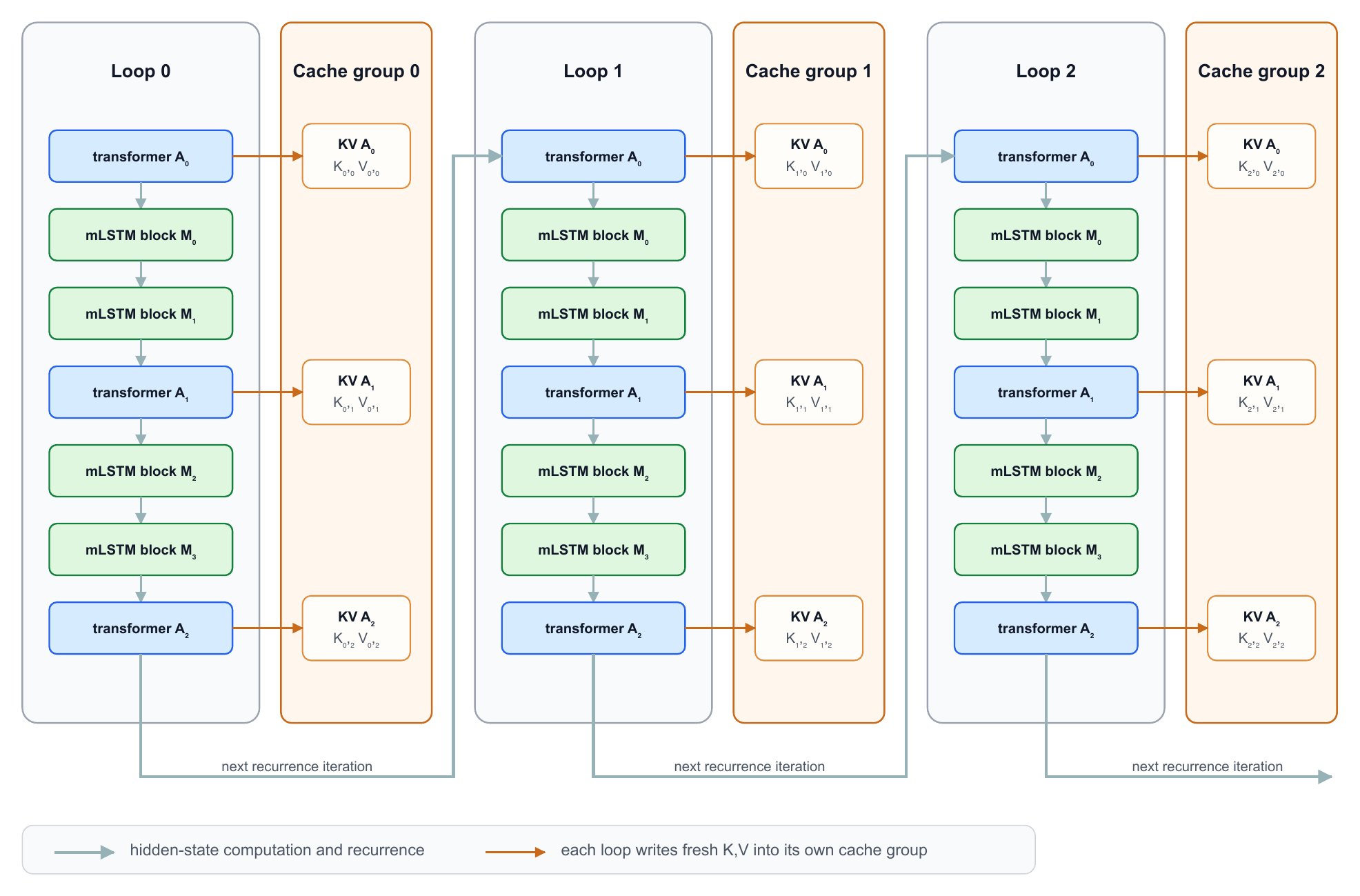}
    \caption{
        Per-iteration KV-cache routing. Every recurrence iteration writes to and reads from an independent cache group.
    }
    \label{fig:cache_per_iteration_appendix}
\end{figure}

\begin{figure}[!htbp]
    \centering
    \includegraphics[width=0.90\linewidth]{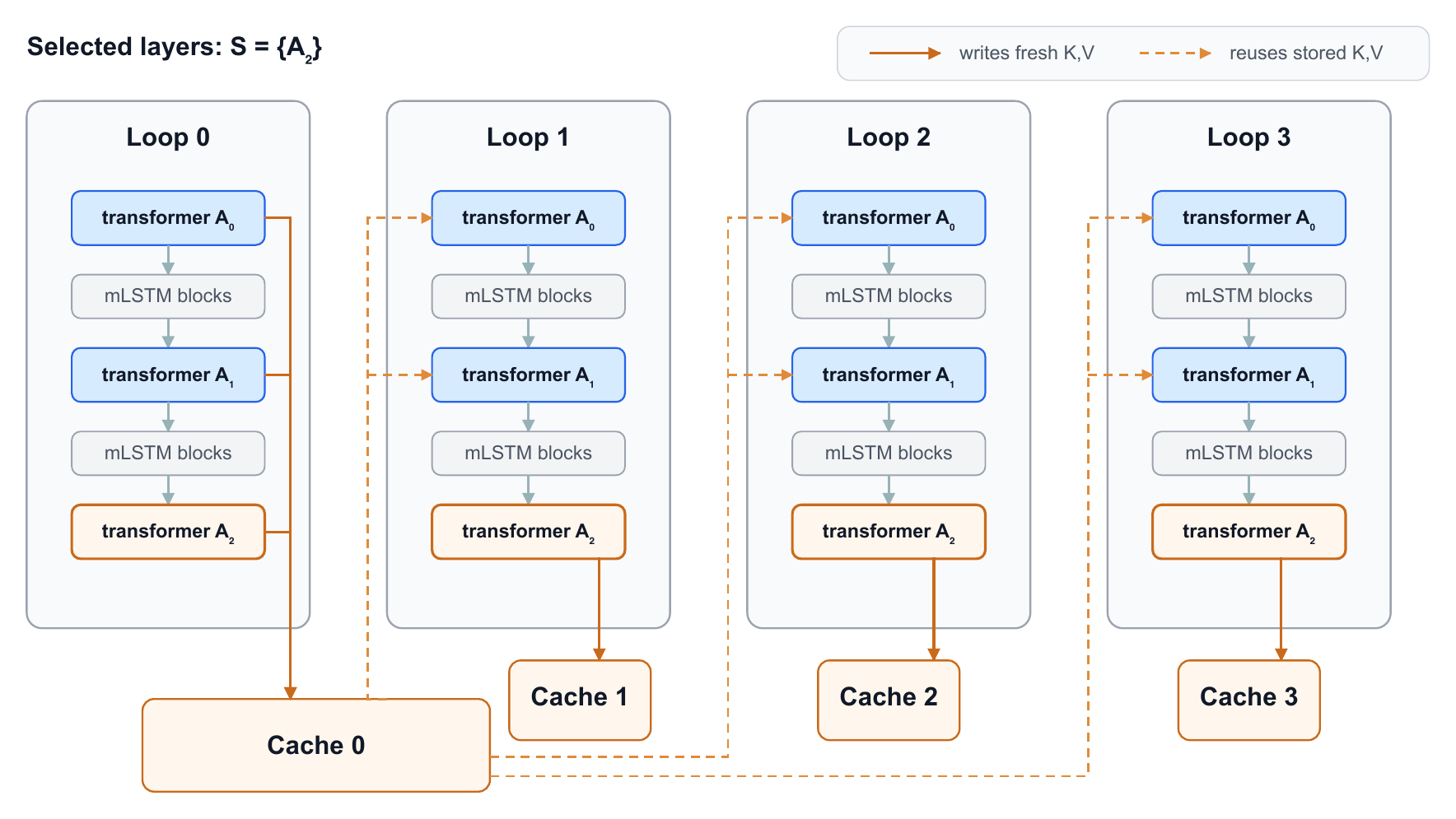}
    \caption{
        Selective-layer KV-cache routing. Selected exact-attention layers retain independent per-iteration caches, while non-selected layers reuse the shared fallback group.
    }
    \label{fig:cache_selective_layer_appendix}
\end{figure}

Figures~\ref{fig:cache_shared_appendix},
\ref{fig:cache_per_iteration_appendix}, and
\ref{fig:cache_selective_layer_appendix} illustrate the cache-routing policies defined in Section~\ref{sec:cache_routing} that are omitted from the main paper. Orange connections denote cache writes, dashed connections denote cache reuse, and grey connections denote hidden-state propagation.

\FloatBarrier

\section{Experimental Setup}
\label{app:setup}

\subsection{Training Data and Evaluation}

All models are trained on the Nemotron-CC-HQ corpus using the
\texttt{m2rnn-400m} tokenizer. Evaluation uses the Hugging Face model interface through \texttt{lm-evaluation-harness}. Unless stated otherwise, reported downstream metrics include PIQA, WinoGrande, ARC-Easy, HellaSwag, ARC-Challenge, SciQ, CORE, CORE Extended, and WikiText perplexity. Models are loaded from the exported Hugging Face checkpoints with batch size \(1\), BOS-token insertion enabled, and chat-template application disabled. The same selected checkpoint is used for all downstream evaluations.

\subsection{Training Configurations}
\paragraph{Main Looped Hybrid.}

Unless stated otherwise, looped-hybrid experiments use 24 logical layers, 4 recurrence iterations, and sequence length 4096. Because recurrent mixer parameters are iteration-specific, parameter count varies with the dense-anchor stride, ranging from \(965.004\)M parameters at \(s_d=2\) to \(1472.851\)M at \(s_d=12\). Exact counts are reported in Appendix~\ref{app:parameter_counts}.

The effective training budget is

\[
4096 \times 64 \times 4 \times 4800
\approx
5.03\times10^9
\]

tokens.

Training uses TorchAdamW with learning rate \(3\times10^{-3}\), weight decay \(0.1\), betas \((0.9,0.95)\), cosine learning-rate decay, and 480 warm-up steps. Models are trained for 4800 optimisation steps on four GPUs using BF16 precision, microbatch size \(1\), gradient accumulation \(64\), global batch size \(256\), and activation checkpointing.

Training and evaluation were performed on the EuroHPC Leonardo Booster supercomputer (4$\times$ NVIDIA A100 64GB GPUs per node) and the EuroHPC Deucalion supercomputer. Unless otherwise stated, the main experiments were trained on 4 GPUs using BF16 precision with activation checkpointing.

\paragraph{Parcae-Matched Configuration.}

The matched configuration follows the original \textsc{Parcae} training setup as closely as practical while using the common training infrastructure. Models use 12 layers, 8 recurrence iterations, sequence length 2048, BF16 precision, learning rate \(2\times10^{-4}\), gradient clipping with maximum norm \(1.0\), no warm-up, and linear cooldown over the final 50\% of training. Unless otherwise stated, the target token budget is approximately \(5.03\times10^9\) tokens.

\paragraph{\textsc{Parcae} with mLSTM.}

Unless otherwise stated, \textsc{Parcae} uses a recurrent workspace dimension of 1024 with a \(4/4/4\) prelude/core/coda decomposition, zero recurrent-state initialisation, disabled prelude normalisation, 8 recurrence iterations, backpropagation depth 4, diagonal input injection, per-sequence recurrence scheduling, Poisson-truncated-full sampling, and lockstep sampling for both recurrence variables \(n\) and \(k\).

\paragraph{Reproducibility.}

Unless otherwise stated, all architectures are trained with a single training seed (42). Checkpoints are selected solely according to the lowest validation loss among the predefined checkpoint saves. Perplexities are computed once from the selected checkpoint and are not averaged across training seeds.
\subsection{Compute Resources}
\label{app:compute}
We conducted the experiments on the EuroHPC Leonardo Booster system at CINECA and the Deucalion GPU partition. On Leonardo Booster nodes we used four NVIDIA A100 GPUs with 64~GB memory each, a 32-core Intel Xeon 8358 CPU, and 512~GB of host memory per node. Deucalion GPU nodes contain four NVIDIA A100 GPUs, a 64-core AMD EPYC 7742 CPU, and 512~GB of host memory. We used approximately 2,400 Leonardo node-hours and 400 Deucalion node-hours, for approximately 2,800 node-hours in total. This included failed runs as well as the runs reported in the paper. Individual training runs took up to 4 days, while evaluation runs took up to 24 hours. Training used BF16 precision and activation checkpointing. The remaining training and evaluation settings are reported in Appendix~\ref{app:setup}.

\section{Implementation Details}
\label{app:implementation}

\subsection{Depth-Recurrence Modes}
\label{app:depth_recurrence_modes}

Our implementation exposes several depth-recurrence modes. Only
\texttt{none}, \texttt{looped\_hybrid}, \texttt{parcae}, and the full looped Transformer configuration are used for the reported experiments.

\paragraph{\texttt{none}.}
A standard non-recurrent Transformer executes the physical stack once.

\paragraph{\texttt{looped\_hybrid}.}
Transformer attention anchors are interleaved with recurrent sequence mixers, and the complete logical stack is repeated. This is the architecture defined in Section~\ref{sec:looped_hybrid}.

\paragraph{\texttt{parcae}.}
Fixed Transformer prelude and coda stages surround an iterated recurrent workspace. This is the architecture defined in Section~\ref{sec:parcae}.

\paragraph{Full looped Transformer.}
The Transformer stack is repeatedly applied with shared parameters. Its cross-iteration KV behaviour is determined by the cache-routing and replay policies defined in Sections~\ref{sec:cache_compression}
and~\ref{sec:cache_routing}.

\paragraph{Additional implementation modes.}
The codebase also supports \texttt{dense}, \texttt{kv\_shared}, \texttt{compressed}, \texttt{compressed\_refresh}, and \texttt{parcae\_cache}. These exploratory modes are not included in the reported experiments.

\subsection{Model Parameter Counts}
\label{app:parameter_counts}

Parameter counts are obtained directly from the exported model state dictionaries. The non-recurrent Transformer baseline and the full looped Transformer share the same \(406.37\)M parameters. The looped-hybrid parameter count depends on the number of recurrent mixer blocks, while the \textsc{Parcae} parameter count depends only on the recurrent workspace dimension \(d_r\).

\begin{table}[t]
    \centering
    \caption{Parameter counts for the reported model configurations.}
    \label{tab:model_parameter_counts}
    \small
    \begin{tabular}{lcc}
        \toprule
        Model & Configuration & Parameters (M) \\
        \midrule
        Transformer baseline & -- & 406.373 \\
        Full looped Transformer & all cache policies & 406.373 \\
        \midrule
        Looped Hybrid mLSTM (main) & \(s_d=2\) & 965.004 \\
        Looped Hybrid mLSTM (main) & \(s_d=3\) & 1168.143 \\
        Looped Hybrid mLSTM (main) & \(s_d=4\) & 1269.712 \\
        Looped Hybrid mLSTM (main) & \(s_d=6\) & 1371.282 \\
        Looped Hybrid mLSTM (main) & \(s_d=8\) & 1422.066 \\
        Looped Hybrid mLSTM (main) & \(s_d=12\) & 1472.851 \\
        Looped Hybrid mLSTM (matched) & \(s_d=3\) & 994.927 \\
        Looped Hybrid mLSTM (matched) & \(s_d=4\) & 1100.692 \\
        Looped Hybrid mLSTM (matched) & \(s_d=6\) & 1206.458 \\
        Looped Hybrid mLSTM + adapter & main, \(s_d=4\) & 1272.073 \\
        Looped Hybrid GDN (main) & main, \(s_d=3\) & 1168.082 \\
        \midrule
        \textsc{Parcae} & \(d_r=768\) & 241.103 \\
        \textsc{Parcae} & \(d_r=1024\) & 259.755 \\
        \textsc{Parcae} & \(d_r=1536\) & 315.935 \\
        \bottomrule
    \end{tabular}
\end{table}

\section{Additional Results}
\label{app:additional_results}

This section provides the complete evaluation results omitted from the main paper, including the full looped-hybrid sweep and the complete \textsc{Parcae} recurrent-workspace sweep.

\subsection{Complete Looped-Hybrid Results}
\label{app:full_results}

Table~\ref{tab:full_results} reports the complete looped-hybrid evaluation results, including the baseline and the perplexity omitted from the main-paper figure.

\begin{table*}[t]
\centering
\caption{Complete evaluation results for the Looped-Hybrid experiments. ``Main'' denotes the \(24\)-layer, four-iteration, \(4096\)-token setup; ``matched'' denotes the \(12\)-layer, 8-iteration, \(2048\)-token setup. We report normalised accuracy on PIQA, WinoGrande, ARC-Easy, HellaSwag, ARC-Challenge, and SciQ; CORE metrics are reported in percentage points. The table includes WikiText perplexity omitted from the main-paper results table.}
\label{tab:full_results}
\scriptsize
\setlength{\tabcolsep}{2.7pt}
\resizebox{\textwidth}{!}{%
\begin{tabular}{lccccccccccccc}
\toprule
Model & \(R\) & \(L\) & \(s_d\) & PIQA-N & Wino. & ARC-E-N & Hella.-N & ARC-C-N & SciQ-N & CORE & CORE Ext. & WikiText PPL \\
\midrule
Transformer baseline
& 1 & 24 & -- & 0.664 & 0.500 & 0.497 & 0.369 & 0.276 & 0.713
& \(14.95\pm0.29\) & \(10.26\pm0.39\) & 35.988 \\

\midrule
mLSTM, main
& 4 & 24 & 2 & 0.633 & 0.505 & 0.448 & 0.368 & 0.257 & 0.641
& \(13.56\pm0.25\) & \(9.87\pm0.35\) & 57.423 \\
mLSTM, main
& 4 & 24 & 3 & 0.644 & 0.522 & 0.447 & 0.372 & 0.268 & 0.674
& \(13.08\pm0.13\) & \(9.12\pm0.19\) & 52.008 \\
mLSTM, main
& 4 & 24 & 4 & 0.652 & 0.521 & 0.473 & 0.372 & 0.266 & 0.677
& \(13.16\pm0.15\) & \(9.35\pm0.38\) & 48.746 \\
mLSTM, main
& 4 & 24 & 6 & 0.649 & 0.511 & 0.447 & 0.353 & 0.270 & 0.626
& \(11.76\pm0.02\) & \(8.20\pm0.09\) & 52.524 \\
mLSTM, main
& 4 & 24 & 8 & 0.587 & 0.506 & 0.391 & 0.275 & 0.229 & 0.490
& \(5.94\pm0.05\) & \(4.31\pm0.11\) & 105.098 \\
mLSTM, main
& 4 & 24 & 12 & 0.583 & 0.511 & 0.386 & 0.271 & 0.224 & 0.500
& \(4.70\pm0.11\) & \(4.80\pm0.14\) & 115.087 \\

\midrule
mLSTM, matched
& 8 & 12 & 3 & 0.626 & 0.500 & 0.422 & 0.308 & 0.265 & 0.573
& \(10.83\pm0.12\) & \(7.99\pm0.31\) & 83.549 \\
mLSTM, matched
& 8 & 12 & 4 & 0.626 & 0.524 & 0.441 & 0.307 & 0.246 & 0.575
& \(9.22\pm0.16\) & \(6.24\pm0.23\) & 82.179 \\
mLSTM, matched
& 8 & 12 & 6 & 0.624 & 0.507 & 0.415 & 0.305 & 0.246 & 0.587
& \(6.89\pm0.11\) & \(5.97\pm0.21\) & 86.494 \\

\midrule
mLSTM + adapter, main
& 4 & 24 & 4 & 0.534 & 0.492 & 0.294 & 0.259 & 0.224 & 0.304
& \(-1.36\pm0.04\) & \(1.49\pm0.09\) & 1324.289 \\
mLSTM + adapter, main
& 4 & 24 & 6 & 0.523 & 0.497 & 0.281 & 0.262 & 0.243 & 0.273
& \(-1.67\pm0.03\) & \(1.41\pm0.12\) & 2767.665 \\
mLSTM + adapter, main
& 4 & 24 & 8 & 0.535 & 0.496 & 0.300 & 0.257 & 0.206 & 0.265
& \(-1.28\pm0.03\) & \(1.21\pm0.04\) & 1666.295 \\
mLSTM + adapter, main
& 4 & 24 & 12 & 0.536 & 0.501 & 0.293 & 0.249 & 0.230 & 0.278
& \(-1.88\pm0.06\) & \(0.96\pm0.06\) & 961.664 \\

\midrule
Gated DeltaNet, main
& 4 & 24 & 2 & 0.669 & 0.524 & 0.530 & 0.430 & 0.305 & 0.771
& \(18.08\pm0.11\) & \(11.77\pm0.17\) & 30.786 \\
Gated DeltaNet, main
& 4 & 24 & 3 & 0.676 & 0.528 & 0.537 & 0.441 & 0.301 & 0.782
& \(\mathbf{18.79\pm0.06}\) & \(12.62\pm0.11\) & 30.703 \\
Gated DeltaNet, main
& 4 & 24 & 4 & \(\mathbf{0.696}\) & 0.522 & \(\mathbf{0.563}\) & 0.446 & 0.292 & \(\mathbf{0.789}\)
& \(16.26\pm0.06\) & \(11.03\pm0.22\) & 30.229 \\
Gated DeltaNet, main
& 4 & 24 & 6 & 0.680 & \(\mathbf{0.531}\) & 0.543 & \(\mathbf{0.449}\) & \(\mathbf{0.315}\) & 0.765
& \(\mathbf{18.79\pm0.11}\) & \(\mathbf{12.70\pm0.09}\) & \(\mathbf{29.974}\) \\
\bottomrule
\end{tabular}%
}
\end{table*}

\subsection{Complete \textsc{Parcae} Results}
\label{app:parcae_results}

Table~\ref{tab:parcae_results_full} reports the complete recurrent-workspace sweep for the matched \(12\)-layer, 8-iteration \textsc{Parcae} setup.

\begin{table*}[t]
    \centering
    \caption{Results of our \textsc{Parcae} implementation under the \(12\)-layer, 8-iteration matched setup. Architecture denotes the prelude/core/coda allocation, \(d_r\) is the recurrent workspace dimension. The architectural sweep uses the common \(5.03\)B-token budget; the final row reports the \(2/4/6,\;d_r=1536\) model trained for the original \textsc{Parcae} budget of \(29.6\)B tokens. The non-recurrent Transformer reference uses the main \(24\)-layer setup. CORE metrics are reported in percentage points.
    }
    \label{tab:parcae_results_full}
    \scriptsize
    \setlength{\tabcolsep}{2.7pt}
    \resizebox{\textwidth}{!}{%
    \begin{tabular}{lccccccccccc}
        \toprule
        Model
        & \(d_r\)
        & Arch.
        & PIQA-N
        & Wino.
        & ARC-E-N
        & Hella.-N
        & ARC-C-N
        & SciQ-N
        & CORE
        & CORE Ext.
        & WikiText PPL \\
        \midrule

        Transformer baseline
        & --
        & \(24/0/0\)
        & \textbf{0.664}
        & 0.500
        & \textbf{0.497}
        & \textbf{0.369}
        & \textbf{0.276}
        & \textbf{0.713}
        & \(\mathbf{14.95\pm0.29}\)
        & \(\mathbf{10.26\pm0.39}\)
        & \textbf{35.988} \\

        \midrule

        \textsc{Parcae} with mLSTM
        & 768
        & \(2/4/6\)
        & 0.634
        & 0.502
        & 0.440
        & 0.313
        & 0.256
        & 0.642
        & \(10.50\pm0.17\)
        & \(7.57\pm0.13\)
        & 48.741 \\

        \textsc{Parcae} with mLSTM
        & 1024
        & \(2/4/6\)
        & 0.632
        & 0.498
        & 0.445
        & 0.315
        & 0.253
        & 0.655
        & \(11.36\pm0.06\)
        & \(8.33\pm0.35\)
        & 47.243 \\

        \textsc{Parcae} with mLSTM
        & 1536
        & \(2/4/6\)
        & 0.632
        & 0.510
        & 0.436
        & 0.323
        & 0.236
        & 0.668
        & \(12.47\pm0.16\)
        & \(8.80\pm0.39\)
        & 45.289 \\

        \midrule

        \textsc{Parcae} with mLSTM
        & 768
        & \(4/4/4\)
        & 0.632
        & \textbf{0.524}
        & 0.439
        & 0.313
        & 0.266
        & 0.631
        & \(12.06\pm0.17\)
        & \(8.90\pm0.23\)
        & 48.700 \\

        \textsc{Parcae} with mLSTM
        & 1024
        & \(4/4/4\)
        & 0.632
        & 0.507
        & 0.445
        & 0.324
        & 0.265
        & 0.631
        & \(11.40\pm0.08\)
        & \(8.21\pm0.16\)
        & 46.651 \\

        \textsc{Parcae} with mLSTM
        & 1536
        & \(4/4/4\)
        & 0.633
        & 0.515
        & 0.449
        & 0.326
        & 0.247
        & 0.650
        & \(12.09\pm0.15\)
        & \(8.42\pm0.11\)
        & 45.460 \\

        \midrule

        \textsc{Parcae} with mLSTM
        & 768
        & \(6/4/2\)
        & 0.633
        & 0.507
        & 0.447
        & 0.312
        & 0.245
        & 0.649
        & \(11.27\pm0.12\)
        & \(8.63\pm0.16\)
        & 48.991 \\

        \textsc{Parcae} with mLSTM
        & 1024
        & \(6/4/2\)
        & 0.638
        & 0.521
        & 0.447
        & 0.319
        & 0.253
        & 0.654
        & \(10.43\pm0.06\)
        & \(7.95\pm0.25\)
        & 46.993 \\

        \textsc{Parcae} with mLSTM
        & 1536
        & \(6/4/2\)
        & 0.633
        & 0.513
        & 0.456
        & 0.331
        & 0.254
        & 0.647
        & \(11.26\pm0.19\)
        & \(7.80\pm0.10\)
        & 45.260 \\

        \midrule
        \multicolumn{12}{c}{Extended training (original \textsc{Parcae} budget)}\\
        \midrule

        \textsc{Parcae} with mLSTM
        & 1536
        & \(2/4/6\)
        & 0.654
        & 0.513
        & 0.492
        & 0.369
        & 0.276
        & 0.719
        & \(15.10\pm0.06\)
        & \(9.52\pm0.25\)
        & 35.750 \\

        \bottomrule
    \end{tabular}%
    }
\end{table*}

\subsection{Existing Assets and Licenses}
\label{app:licenses}

We use the Nemotron-CC-HQ training corpus, governed by the NVIDIA Data Agreement for Model Training. Our implementation is based on \texttt{lm-engine} at commit \texttt{661c45ce887afcb14b59f41ced3f55b70d435765} and \texttt{xlstm-flash-linear-attention} at commit \texttt{072988b1f23b6702e272069217a7c407a775ff7a}. Evaluation uses \texttt{lm-evaluation-harness} at commit \texttt{64982776449fc5687b6d510e77f5fe6a4a5327ff}, released under the MIT License. The existing methods and models used in the study are cited in the main paper.

\end{document}